\documentclass[lettersize,journal]{IEEEtran}
\usepackage{multirow}
\usepackage{amsmath}
\usepackage{amsfonts}
\usepackage{amssymb}
\usepackage{array}
\usepackage[caption=false,font=normalsize,labelfont=sf,textfont=sf]{subfig}
\usepackage{textcomp}
\usepackage{stfloats}
\usepackage{url}
\usepackage{verbatim}
\usepackage{graphicx}
\usepackage{color}
\usepackage{cite}
\usepackage{bbding}
\usepackage{amssymb}
\usepackage{algpseudocode}
\usepackage{algorithmicx}
\usepackage{algpseudocode}
\usepackage{algorithm}
\usepackage{booktabs}
\DeclareUnicodeCharacter{FF1A}{\textasciitilde}
\usepackage{makecell}
\usepackage[colorlinks,
            linkcolor=red,
            anchorcolor=blue,
            citecolor=green 
            ]{hyperref}

\begin{document}

\title{SIRST-5K: Exploring Massive Negatives Synthesis with Self-supervised Learning for Robust Infrared Small Target Detection }

\author{Yahao Lu, Yupei Lin, Han Wu, Xiaoyu Xian, Yukai Shi\dag, Liang Lin,~\IEEEmembership{Fellow,~IEEE,}
\thanks{
\dag Corresponding author: Yukai Shi.
}
}


\maketitle

\begin{abstract}
Single-frame infrared small target (SIRST) detection aims to recognize small targets from clutter backgrounds. Recently, convolutional neural networks have achieved significant advantages in general object detection. With the development of Transformer, the scale of SIRST models is constantly increasing. Due to the limited training samples, performance has not been improved accordingly. The quality, quantity, and diversity of the infrared dataset are critical to the detection of small targets. To highlight this issue, we propose a negative sample augmentation method in this paper. Specifically, a negative augmentation approach is proposed to generate massive negatives for self-supervised learning. Firstly, we perform a sequential noise modeling technology to generate realistic infrared data. Secondly, we fuse the extracted noise with the original data to facilitate diversity and fidelity in the generated data. Lastly, we proposed a negative augmentation strategy to enrich diversity as well as maintain semantic invariance. The proposed algorithm produces a synthetic SIRST-5K dataset, which contains massive pseudo-data and corresponding labels. With a rich diversity of infrared small target data, our algorithm significantly improves the model performance and convergence speed. Compared with other state-of-the-art (SOTA) methods, our method achieves outstanding performance in terms of probability of detection ($P_d$), false-alarm rate ($F_a$), and intersection over union ($IoU$). The code is available at: \url{https://github.com/luy0222/SIRST-5K}.
\end{abstract}


\section{Introduction}
\IEEEPARstart{I}{nfrared} -small target detection technology~\cite{remotesensing},~\cite{remotesensing4} is a combination of computer vision and infrared remote sensing technology~\cite{environmental1},~\cite{environmental2},~\cite{environmental3},~\cite{TGRS},~\cite{TGRS1}. The advantage of infrared small target detection technology lies in its excellent detection capability in nighttime and adverse weather conditions. Therefore, it has important application value in the fields of remote sensing ~\cite{remotesensing1},~\cite{remotesensing2},~\cite{remotesensing3}, security~\cite{security},~\cite{security1}, and environmental monitoring~\cite{environmental}. 
Due to the long imaging distance, poor visual features, and sensor noise of infrared stream, small target detection has become one of the most challenging tasks in remote sensing and computer vision.

\begin{figure}[ht]
    \centering
    \includegraphics[width=0.45\textwidth]{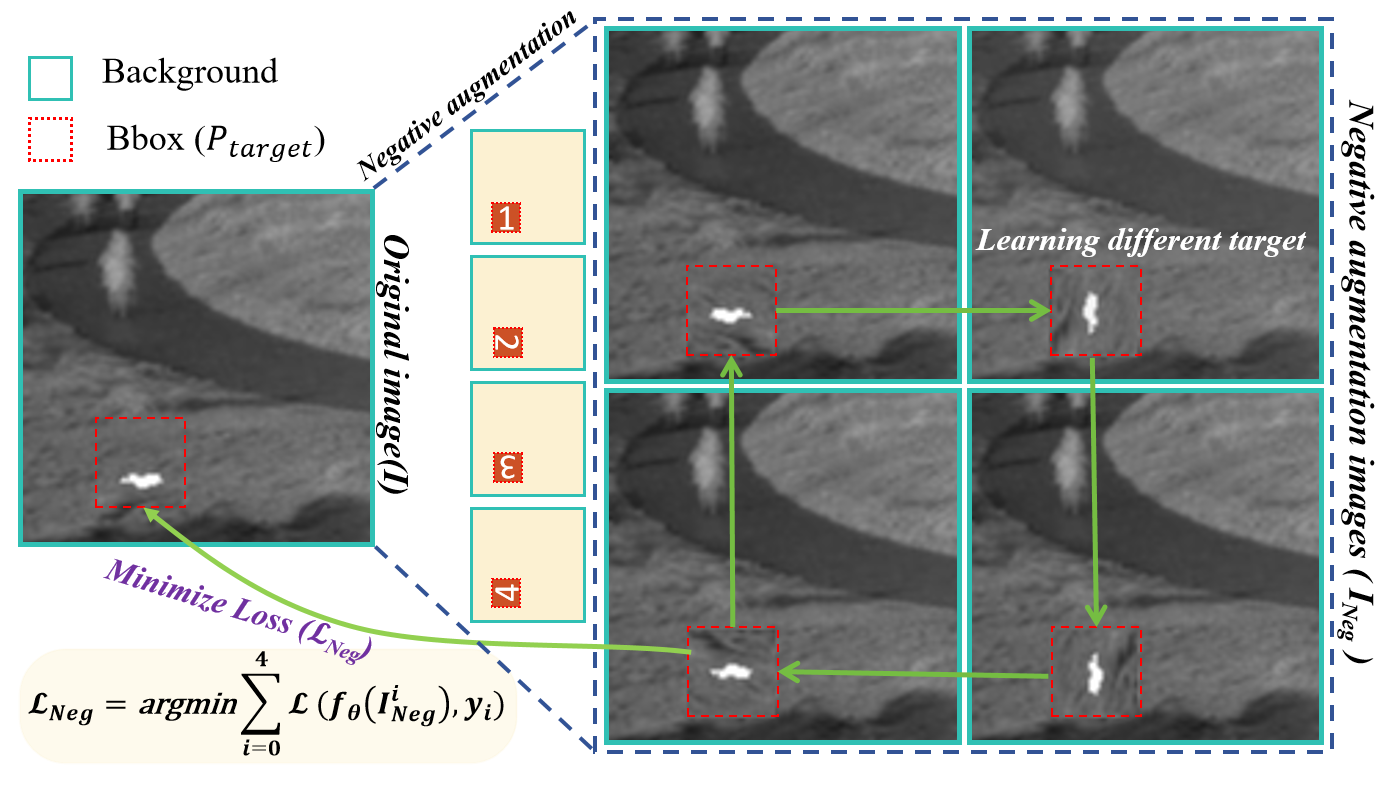}
    \caption[something short]{Self-supervised strategy based on negative augmentation. To explore target correspondence in SIRST, we propose a negative augmentation approach to generate massive negatives for self-supervised representation learning. }
    \label{fig:self-supervised}
\end{figure}

In recent years, with the continuous development of deep learning technology, many researchers have begun to apply it to infrared small target detection~\cite{Deeplearning},~\cite{Deeplearning1}. By using a convolutional neural network (CNN) to learn discriminative feature representations~\cite{cnn}, the accuracy and robustness of infrared small target detection were greatly improved. Liu et al.~\cite{MLP} designed a 5-layer multilayer perceptron (MLP) network for infrared small target detection. As single-frame infrared images lack effective information such as shape and texture, making them difficult to detect effectively. To address this issue, Dai et al.~\cite{ALC} proposed attentional local contrast networks (ALC). By capturing local contrast and integrating the low-level features~\cite{lcm}, local and global information are fused to improve the performance of small target detection. Dai et al.~\cite{ACM} designed an asymmetric context module (ACM) to replace the simple skip connection of U-Net~\cite{unet}. Li et al.~\cite{DNA} proposed a dense nested attention network (DNA) that fully utilizes small target contextual information through repeated fusion, achieving better performance. To further fuse the local features of small targets, Wu et al.~\cite{UIU} proposed a U-Net in U-Net model. They used a new dual-path structure to fuse local and global contextual information together, achieving outstanding small target detection results. 

With the development of Transformer~\cite{vaswani2017attention}, the size of infrared small target detection models is constantly increasing, and their performance has not been improved accordingly. Currently, the scale of infrared small target detection datasets is limited, which restricts the performance of the models. In addition, expanding current datasets is usually expensive and time-consuming. In the past, data augmentation methods have been used to address this problem. However, traditional data augmentation methods, such as flipping, cropping, and rotation, may cause small targets to be warped, causing the absence of labels. 

\begin{figure*}[!t]
\centering
\includegraphics[width=0.99\linewidth]{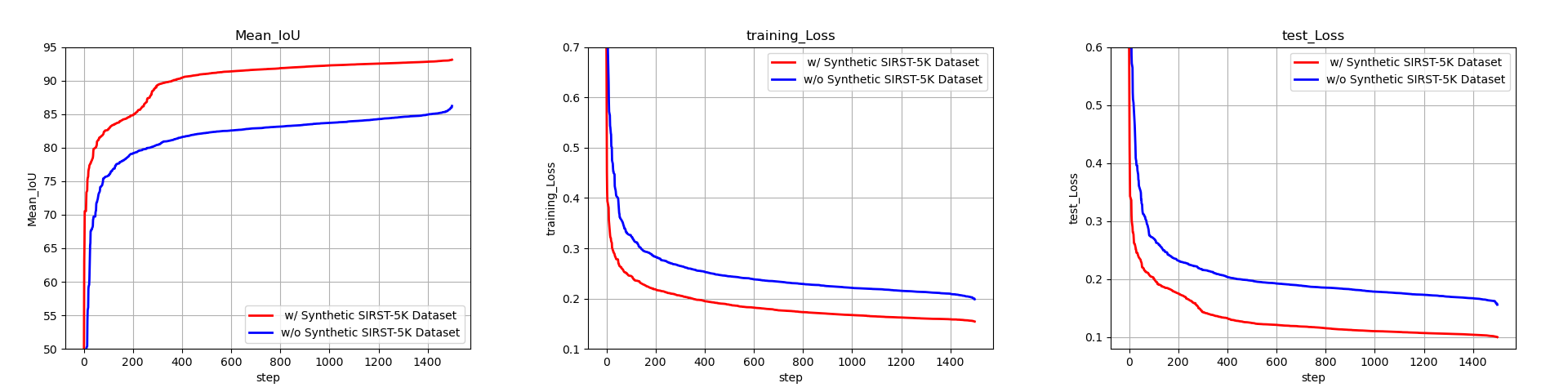}
\caption{The quality, quantity, and diversity of the infrared data have a significant impact on the detection of small targets. By applying a self-supervised learning paradigm with massive pseudo-data, our negative generation strategy has achieved \textbf{faster convergence rate, less training loss and better mean $IoU$.}}
\label{fig2}
\end{figure*}

\emph{The quality, quantity, and diversity of the infrared data have a significant impact on the detection of small targets.} To address the forementioned concerns, we constructed a new SIRST-5K dataset, which contains 5562 images, to improve the quality of the infrared data. In our framework, we provide a new self-supervised learning paradigm with data generation for infrared small target detection. Specifically, a negative augmentation approach is proposed to generate massive negatives for self-supervised learning. Firstly, we perform sequential noise modeling technology to generate realistic infrared data. Secondly, we fuse the extracted noise with the original data to facilitate diversity and fidelity in the generated data. Then, we proposed a negative augmentation strategy to enrich diversity as well as maintain semantic invariance. As shown in Fig.~\ref{fig1} and~\ref{fig2}, given limited infrared small target data, our algorithm has achieved faster convergence and better performance. In summary, the contributions of this paper can be summarized as follows:
\begin{itemize}
    
    \item[$\bullet$] Limited infrared images are renewed with richer diversity through the proposed negative generation strategy. By applying a self-supervised learning paradigm with massive pseudo-data, our model exhibits faster convergence, better performance and strong versatility. 
    
    \item[$\bullet$] We utilize a Noise2Noise displacement strategy to ensure the fidelity and diversity of the synthesized infrared data. Without any manual annotation, the Noise2Noise displacement strategy effectively improves the accuracy and robustness of infrared small target detection. 
    
    \item[$\bullet$]Extensive experiments demonstrate that our method has achieved state-of-the-art results. Compared to other state-of-the-art methods, our approach yields outstanding improvements of $P_d$, $F_a$, and $IoU$.
\end{itemize}

\section{RELATED WORK}
\subsection{Single-frame Infrared Small Target Detection}
Single-frame infrared small target (SIRST) detection is an important problem in infrared image processing. The goal of SIRST is to accurately detect the target of interest in a single infrared image. Due to the diverse characteristics of infrared images such as low contrast and noise interference, single-frame infrared small target detection faces great challenges.

\begin{figure}[t]
\centering
\includegraphics[width=0.99\columnwidth]{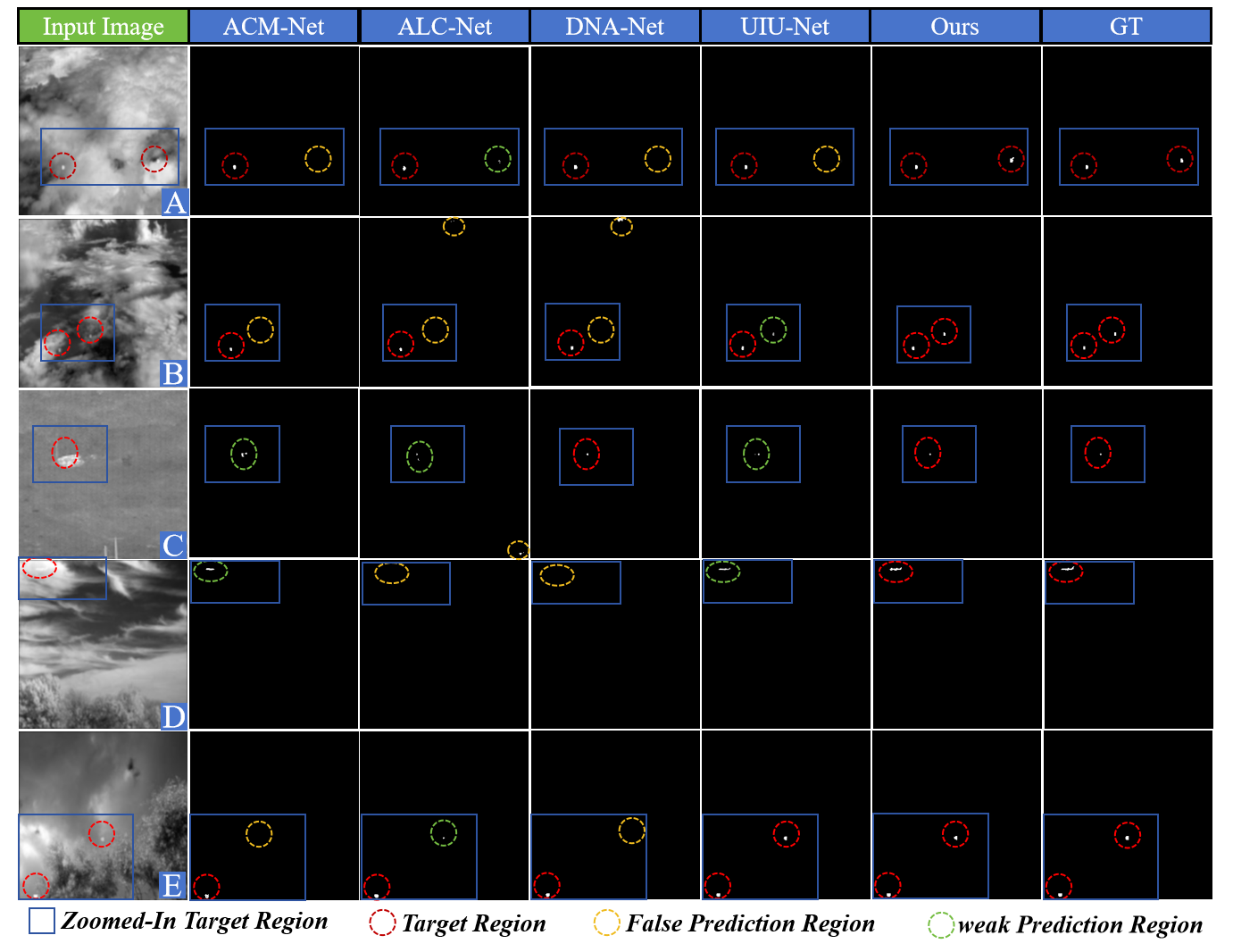}
\caption{The qualitative results of different SIRST detection methods. The correctly detected targets, false alarms, and weak detection regions are highlighted using \textcolor{red}{red}, \textcolor[rgb]{0.75, 0.5, 0.25}{yellow}, and \textcolor{green}{green} dashed circles respectively. Our method performs accurate target localization with a lower false alarm rate.}
\label{fig1}
\end{figure}

Traditional methods based on feature extraction, including filter-based and morphological methods. These methods manually design features to extract useful information and then use classifiers for target detection. Considering the directionality of background clutter. Li et al.~\cite{MITHF} developed a multi-directional improved top hat filter (MITHF) to further enhance clutter suppression. Bai et al~\cite{Bai} constructed a new transform by modifying the operation rules and structural elements in mathematical morphology filtering, which has good background estimation capabilities. However, traditional methods require manual feature design and are sensitive to complex backgrounds and noise interference, which have certain limitations in practical applications. In order to improve the effectiveness and robustness of feature extraction, researchers are exploring solid feature extraction methods, such as convolutional neural networks.

Targets in single-frame infrared images are usually very small and lack effective information such as shape and texture, making effective detection difficult. McIntosh et al.~\cite{Mclntosh}  fine-tuned Faster-RCNN~\cite{rcnn} and Yolo-v3~\cite{yolov3}, using optimized feature vectors as input to achieve improved performance. Wang et al.~\cite{Bai} proposed a conditional generative adversarial network (GAN) consisting of two generators and one discriminator, achieving low rates of missed detection and false alarm ($F_a$) through adversarial training. Dai et al~\cite{ALC}. proposed attentional local contrast networks (ALC). By capturing local contrast information and integrating the detailed information of low-level features into high-level features, ALC-Net achieves excellent small target detection. Xi et al.~\cite{xi} proposed the Infrared Small Target Detection U-Net (ISTDU-Net), converting single-frame infrared images into probability likelihood maps of infrared small targets. A fully connected layer was introduced in the neural network to improve the contrast between targets and backgrounds. Dai et al.~\cite{ACM} designed an asymmetric context module (ACM) to replace the simple skip connection of U-Net. Li et al.~\cite{DNA} proposed the densely nested attention network (DNA-Net), by repeatedly fusing and enhancing small target context information through dense nesting. To make the local features of small targets more fully fused, Wu et al.~\cite{UIU} proposed an infrared small target detection model called U-Net in U-Net , which uses a new dual-path structure to fuse local context information and global context information together, achieving excellent small target detection performance.

Although the performance of infrared small target detection networks has been significantly improved. Small targets in single-frame infrared images are usually rare, resulting in limited data sources. 

\subsection{Data Augmentation for Representation Learning}
Data augmentation refers to the process of increasing the amount of data by creating synthetic data from existing data. However, basic data augmentation methods may alter the semantic information of images, leading to a decrease in image quality and overfitting issues. 

\begin{figure*}[!t]
\centering
\includegraphics[width=0.95\linewidth]{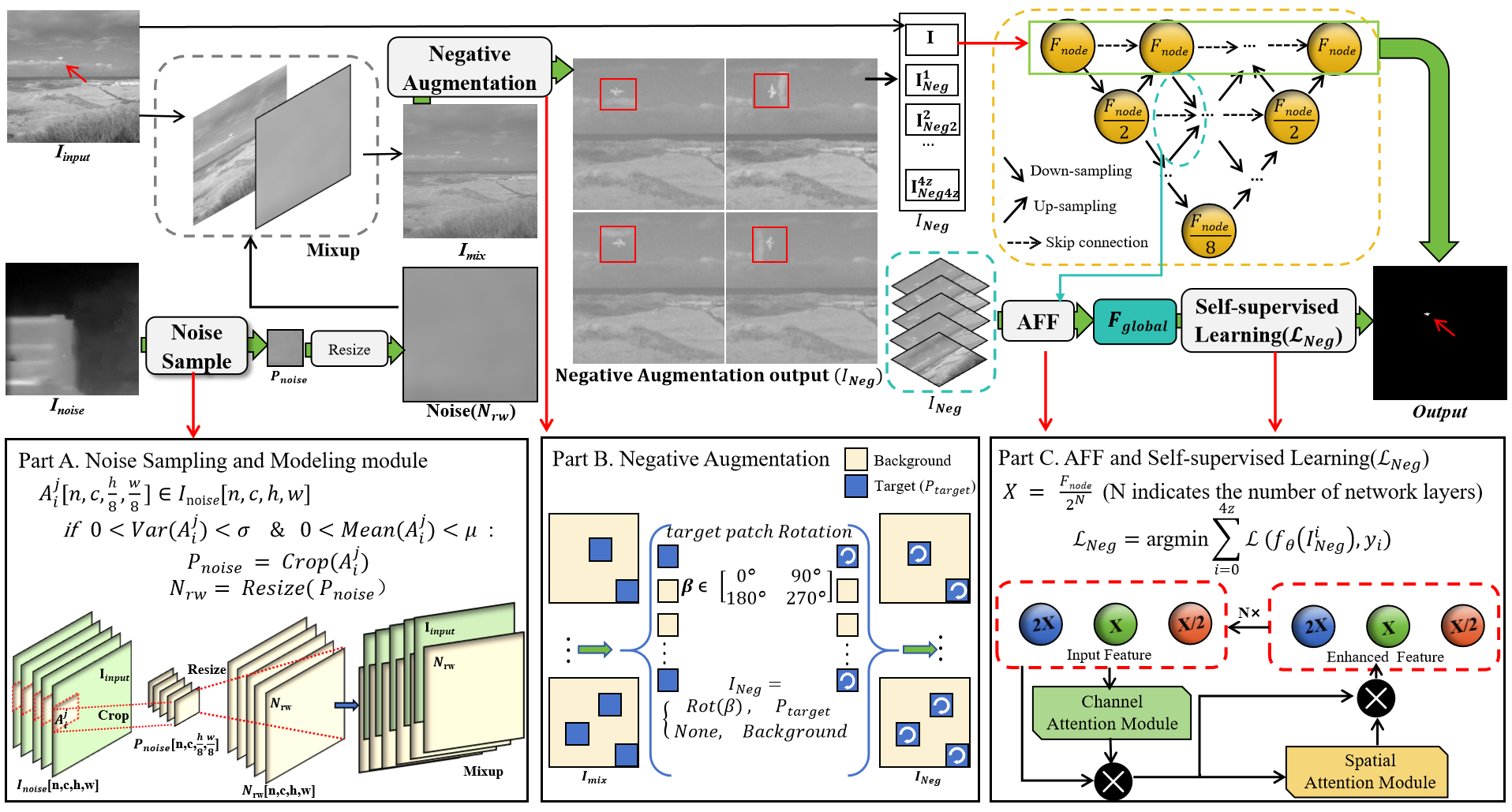}
\caption{Illustration of the proposed massive negatives synthesis framework.  (a) Noise Sampling and Modeling module. The input image is first divided into equally sized local regions. Qualified noise sampling regions are selected and resized to fetch real-world noise. Then, the training samples are mixed with real-world noise; (b) Negative Augmentation module. Among the generated samples, the challenging small targets are further processed with negative augmentation to produce massive negatives; (c) Self-supervised Learning. By utilizing these negative samples and their corresponding labels, we can implement self-supervised representation learning to learn richer feature representations.}
\label{fig4}
\end{figure*}

Recently, Mixup~\cite{mixup} and its derivatives have achieved excellent results in many tasks. Mixup generates new images by linearly interpolating and mixing the pixel values of two different images using weights. This method can improve the generalization ability and robustness of models. With the introduction of Mixup, many improved methods based on Mixup have emerged. CutMix~\cite{cutmix} adopts the form of cropping and patching parts of an image to mix it, and the mixing position uses a binarized mask. Saliencymix~\cite{saliencymix} considers that different regions contain varying amounts of information when performing mixing, selecting targeted regions for mixing. Co-mixup~\cite{comixup} transforms the search for the optimal mixup method into an optimization problem by introducing multiple mix targets, guiding the mixup process. Co-mixup also employs saliency guidance, extracting salient regions from multiple samples and mixing them. By quantifying saliency and designing a submodular minimization algorithm, it aims to maximize the cumulative salient regions in generated images. Alignmix~\cite{alignmixup} further addresses the quality issue of fused images, proposing to align each position of two images in the feature space before performing difference operations at corresponding positions, resulting in higher-quality image fusion. Stylemix~\cite{stylemix} focuses on decomposing the style and content of an image separately for mixing, aiming to achieve higher-quality mixed image generation. TokenMixup~\cite{tokenmixup} primarily utilizes the attention mechanism inherent in Transformers to calculate the saliency of each image token, maximizing the overall saliency of a batch of mixed data. Mixup and its derivatives can be more effectively used for data augmentation tasks. However, they struggle with handling sample imbalance issues related to small targets. Since small targets naturally have fewer instances, applying Mixup will exacerbate the sample imbalance problem, causing models to be biased towards recognizing small targets.

In infrared small target detection, targets possess non-prominent attributes. For instance, small physical size and weak intensity make them difficult to detect effectively. To emphasize this issue, we propose a negative sample generation technique for handling sparse data. By leveraging negative sample enhancement strategies, we can generate a substantial amount of pseudo data along with corresponding labels, enabling self-supervised representation learning.

\section{Methodology}
In this section, we first provide a framework overview. Then, we delve into the details of the three sub-modular structures, highlighting our key innovations. As shown in Fig.~\ref{fig4}, Our framework comprises three modules: real-world Noise2Noise displacement, negative augmentation, attention feature fusion and self-supervised learning.

\subsection{Real-world Noise2Noise Displacement} 
\textbf {Real-world Noise Sampling: }  We propose a noise sampling method for extracting sequence noise from a dataset and performing noise displacement for realistic data synthesis. For the small target detection dataset, most of the data comes from the same infrared sensor, resulting in a unique noise distribution. To enhance data diversity, we introduce the Noise2Noise displacement strategy. Transferring the unique noise distribution to different data, our model can learn more discriminative features. The algorithm of the solving process is shown as Algorithm~\ref{alg:real_image_editing}. Assuming we select $n$ training images as noise samples: $I_{noise} = [I_{noise}^{1},I_{noise}^{2},...,I_{noise}^n]\in R^{n \times c \times h \times w}$.

Given a noise sample from $I_{noise}$, each noise sampling region ($A_{i}$) has a size of $R^{c\times \frac{h}{8 } \times \frac{w}{8 }}$. Here, $A_{i}$ represents the i-th noise sampling region in our noise sampling image $I_{noise}$, where $i\in [1,64]$. 

\begin{figure}[t]
\centering
\includegraphics[width=0.99\columnwidth]{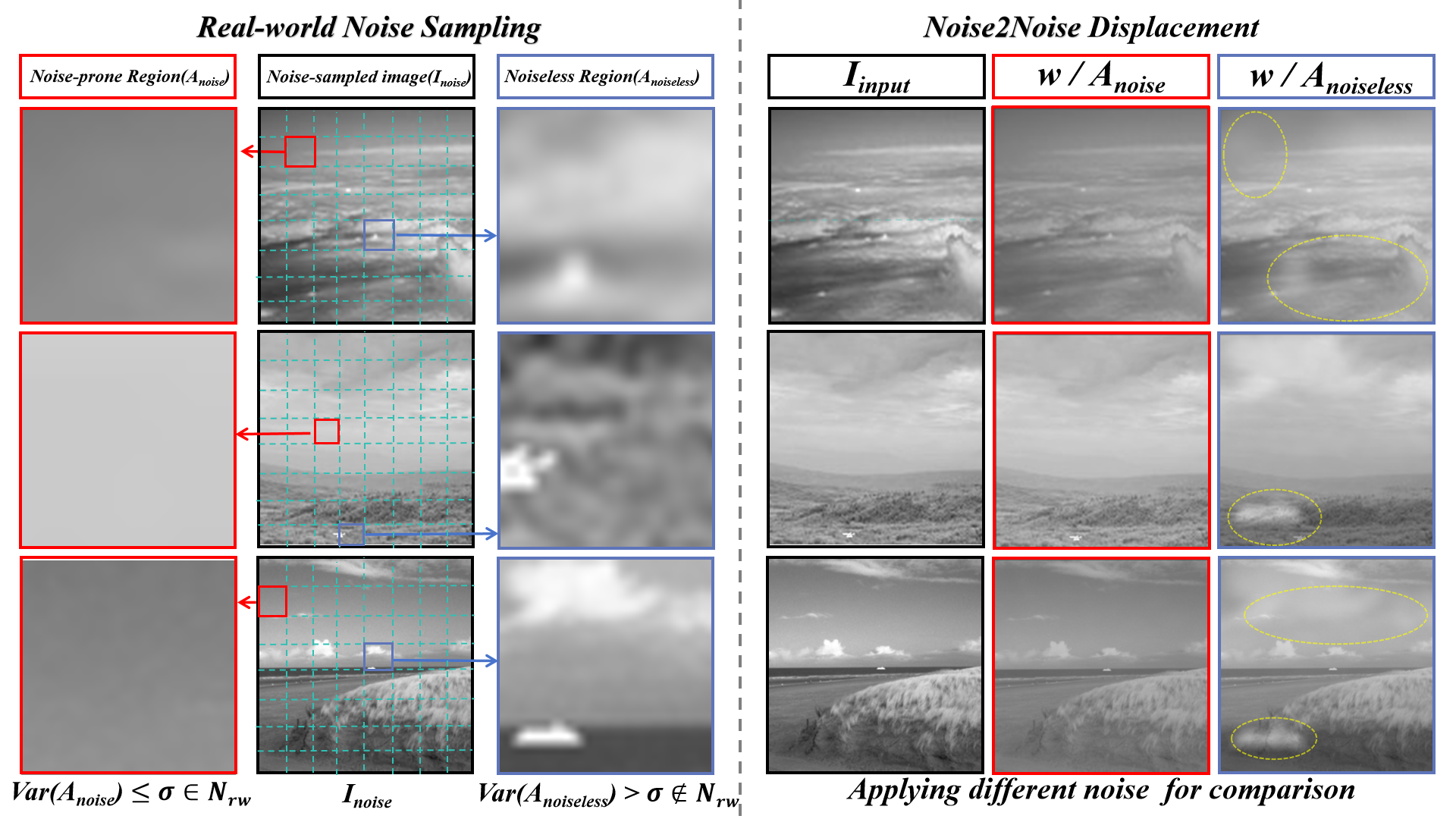}
\caption{A demonstration of noise sampling in infrared small target detection dataset. High-variance noise can affect the model's ability to recognize small targets. The yellow dashed circle highlights the introduced texture. By selecting Noise-prone Region ($A_{noise}$), our framework can fetch diverse noise in infrared sensors to facilitate realistic sample augmentation in SIRST-5K.} 
\label{fig5}
\end{figure}

Usually, noise sampling regions with high variance contain rich textures, which is often cover the noise itself and commonly referred to the noiseless region ($A_{noiseless}$). Conversely, the plain area with low gradient variance has a stronger characteristic in the noise itself. This low gradient variance area is termed the noise-prone region ($A_{noise}$)~\cite{Ji_2020_CVPR_Workshops}. To ensure that the textures in the extracted noise sequence are as uniform as possible, we select noise sampling regions ($A_{i}$), As shown in Fig.~\ref{fig5}. This equation can be summarized as follows:
 \begin{equation}
    P_{noise} = \left\{\begin{matrix} Crop(A_{i}) ,0<Var(A_{i})<\sigma \& 0<Mean(A_{i})<\mu,  \\  None,   otherwise \end{matrix}\right. 
\end{equation}
where $Var( \cdot  )$ and $Mean( \cdot  )$ refer to functions used for calculating variance and mean, respectively. $\sigma$ and $\mu $ represent the maximum variance and maximum mean, respectively. $P_{noise}$ denotes the noise characteristic obtained by the noise-prone region. The noise sampling region that satisfies the constraint conditions is considered as the noise-prone region. We resize the noise characteristic to have the equal size as the noise-sampled image $I_{input}$ in order to generate real-world noise ($N_{rw}$) :
\begin{equation}
   N_{rw} = Resize(P_{noise}),
\end{equation}
where $N_{rw}$ represents real-world noise, and $P_{noise}$ denotes noise-prone region patches. The $Resize( \cdot  )$ indicates resizing the noise-prone to match the dimensions of $I_{input}$. By performing the operations in part A of Fig.~\ref{fig4}, we perform a real-world noise sampling for the infrared sensor data.

\textbf{Noise2Noise Displacement:} 
As shown in Fig.~\ref{fig5}, by adding noise2noise displacement, we can transform the noise in infrared images into more diverse forms. This allows the small target detection model to be exposed to richer noise samples during training. This diverse noise training helps the model learn to accurately identify and locate small targets under different noise conditions. We obtain the auxiliary training samples by displacing the input training image $I_{input}$ as:
\begin{equation} \label{eq1}
   I_{mix} = \alpha  \cdot  N_{rw}+(1-\alpha)  \cdot  I_{input}  ,
\end{equation}
where $I_{mix}\in R^{n \times c \times h \times w} $ represents the training samples that are being mixed with noise, while $I_{input}$  denotes the initially input image, $\alpha \in [0,1]$ indicates the hyperparameter for noise displacement.

\begin{figure}[t]
\centering
\includegraphics[width=0.95\linewidth]{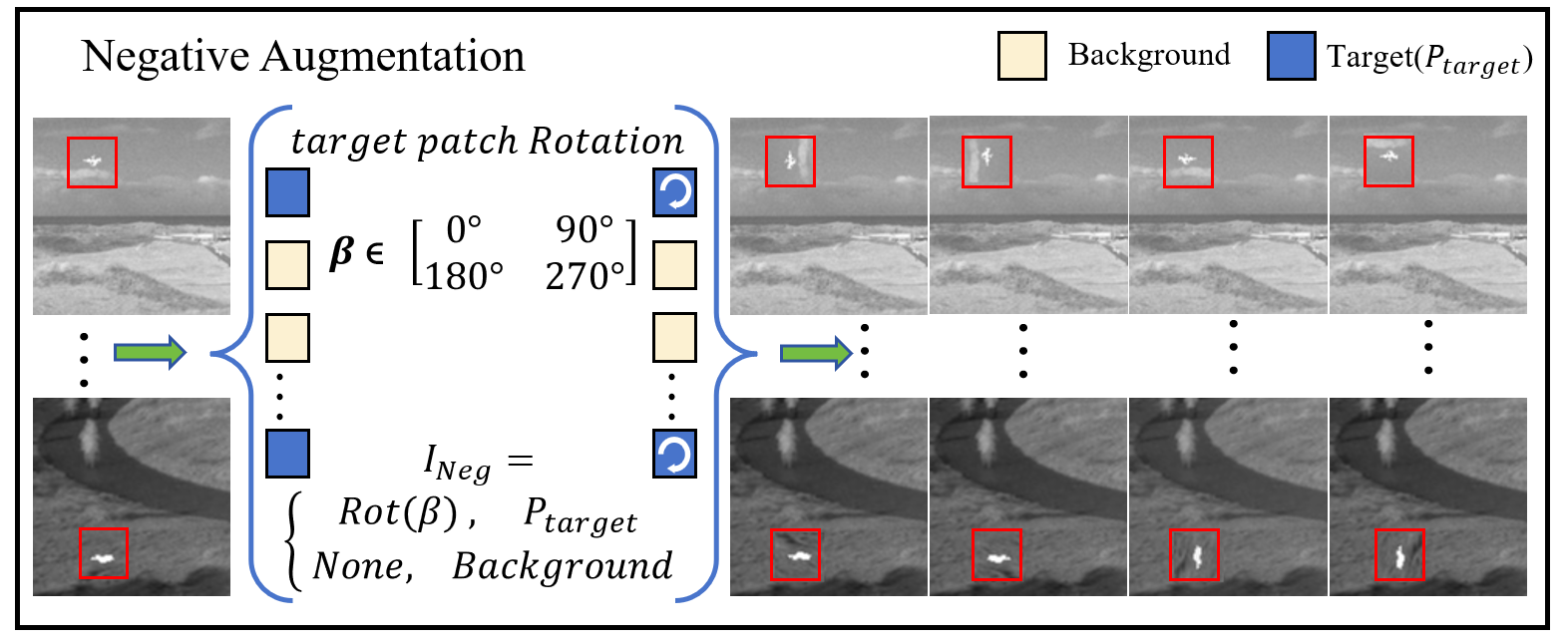}
\caption{With the negative augmentation, limited infrared images are renewed with richer diversity. To make the discriminative model learn effective features on small targets, a negative augmentation strategy is developed for the special yet challenging target augmentation. }
\label{fig6}
\end{figure}

\subsection{Negative Augmentation} \label{NA}
In infrared small target detection, small targets are prone to be ignored or misidentified. To emphasize this issue, we propose a negative sample augmentation strategy to enhance the discriminative feature representation learning. By using the negative sample augmentation strategy, the semantics remain unchanged and the model itself can learn richer features. Specifically, we select an image $I_{mix}$ to demonstrate the negative sample augmentation process, where $I_{mix}$ represents an image mixed with real-world noise ($N_{rw}$). Suppose the anchor region of small targets in the image be denoted as $P_{target} \in[c,s,s]$. Meanwhile, $s$ represents the length and width of $P_{target}$, and $c$ represents the number of channels in $P_{target}$. As shown in Fig.~\ref{fig6}, we perform central rotation on $P_{target}$ to generate negative samples, thereby helping the model learn more diverse features from challenging negatives. This equation can be summarized as follows:
\begin{equation}
   I_{Neg} = Rot(P_{target}, \beta), \beta \in[0^{ \circ }, 90^{ \circ }, 180^{ \circ }, 270^{ \circ }]  ,
\end{equation}
where $I_{Neg}=[I^{1}_{Neg},I^{2}_{Neg},I^{3}_{Neg},I^{4}_{Neg}]\in R^{4 \times c\times h\times w}$ represents the samples after performing negative sample enhancement. $P_{target}$ denotes the target-centered $c \times s \times s$ patch, and $\mu$ represents the rotation angle. $Rot( \cdot  )$ stands for the random central rotation operation. Specifically, if there are $z$ targets within an image ($z > 2$), then the corresponding generated samples $I_{Neg}= [I^{1}_{Neg},I^{2}_{Neg},...,I^{4 \times z}_{Neg}]\in R^{4z \times c \times h \times w}$.

\begin{figure}[t]
\centering
\includegraphics[width=0.99\columnwidth]{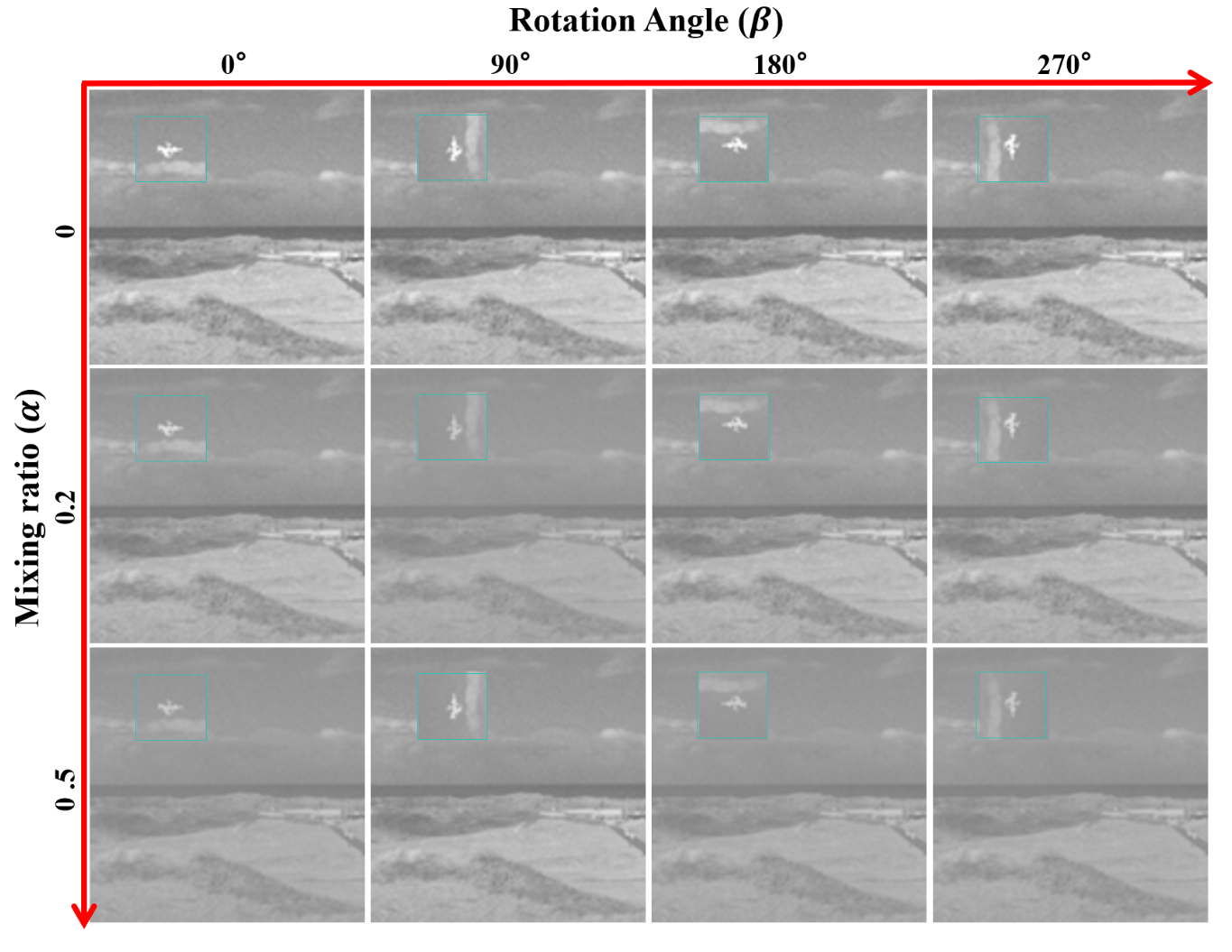}
\caption{Visualization of the samples in SIRST-5K. Among our framework, Real-world Noise2Noise Displacement and Negative Augmentation methods adjust the noise weight and rotation (angle) of the generated image for self-supervised learning. }
\label{fig3}
\end{figure}

\subsection{Attention Feature Fusion (AFF)}
\begin{figure}[t]
\centering
\includegraphics[width=0.99\columnwidth]{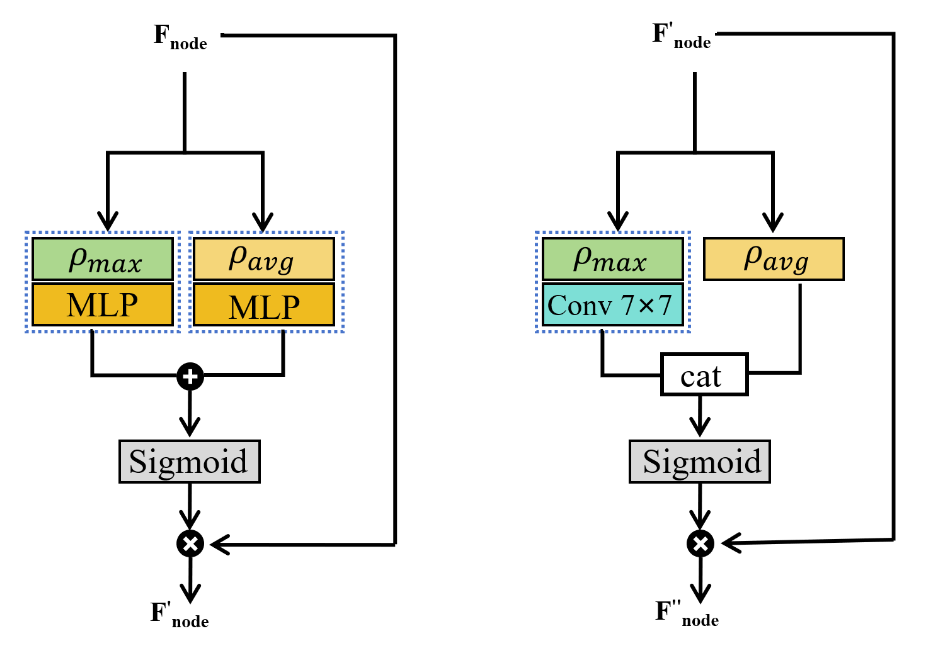}
\caption{Attention Feature Fusion module is used to reduce the semantic gap during the multi-layer feature fusion stage. This diagram represents \textbf{Channel Attention Module (left)} and \textbf{Spatial Attention Module (right)}, respectively}
\label{fig7}
\end{figure}

We propose a network model $f_{ \theta }$ based on the U-Net architecture to extract features for infrared small target images. Our network $f_{ \theta }$ consists of an encoder, a decoder, and skip connections. To better fuse features, we apply the channel attention model and spatial attention model, As shown in Fig.~\ref{fig7}. The features of any feature node $F_{node}$ in ours $f_{ \theta }$ are fused as follows:
\begin{equation}
  F_{Node}' = \sigma [MLP(\rho _{max}(F_{Node})+MLP(\rho _{avg}(F_{Node}))]\odot F_{Node},
\label{eqa:AFF1}
\end{equation}
\begin{equation}
 F_{Node}'' = \sigma [Conv(\rho _{max}(F_{Node}')),(\rho _{avg}(F_{Node}'))]\odot F_{Node}',
\label{eqa:AFF2}
\end{equation}

where $\odot$ represents element-wise multiplication, $F_{Node}$ denotes the feature nodes, and $\sigma$ stands for the sigmoid function. $\rho _{max}$ and $\rho _{avg}$ respectively represent max pooling and average pooling, while $Conv$  is a $7 \times 7$ convolutional layer. The shared network is composed of a multi-layer perceptron (MLP) with one hidden
layer. Equ.~\ref{eqa:AFF1} and~\ref{eqa:AFF2} represent the channel and spatial attention mechanism, respectively.

We utilize an attention mechanism network $f_{ \theta}$ to connect shallow features with rich spatial and contour information and deep features with abundant semantic information, generating a global robust feature map 
$F_{global}$. The network architecture in this paper consists of 4 layers:
\begin{equation}
F_{global} = \left \{F_{Node}'',\frac{F_{Node}''}{2}, \frac{F_{Node}''}{4},\frac{F_{Node}''}{8} \right \} ,
\end{equation}
We apply a loss function for minimizing Soft-IoU is:
\begin{equation}
\mathcal{L}(f_{ \theta }(I_{i}),y_i)=1-\frac{(\sigma (f_{ \theta }(I_{i}))\cap  y_i)}{(\sigma (f_{ \theta }(I_{i}))\cup   y_i)} ,
\end{equation}
where ($I_{i}, y_{i}$) indicates input sample and corresponding annotation, $\mathcal{L}(f_{ \theta }(I_{i}),y_i)$ represents the Soft-IoU loss of $f_{ \theta}$, $f_{ \theta }(I_{i})$ is the predicted result, $\sigma$ signifies the sigmoid function. 

\begin{algorithm}[t!]
\caption{Negative Augmentation}
\label{alg:real_image_editing}
\begin{algorithmic}

\State \textbf{Input:} $I_{noise}\in R^{n\times c\times h\times w} $ : noise sampling images
{$I_{input}\in R^{n\times c\times h\times w}$: input training images}
\State \textbf{Intermediate parameter:} 
\State $A_{i}^{j}$: the i-th noise sampling region of the j-th image
\State  $P_{noise}$: noise-prone area        \\
  \textbf {Output:} $I_{mix}$: \text{ training samples with noise mixing}\\ $I_{Neg}$: \text{ training samples with negative augmentation}

\\ \\
 $\triangleright$ Real-world noise sampling 

\For{$i\gets1$ to K, $j \gets1$ to n}
    \State Var = $Var(A_{i}^{j} )$
    \State Mean = $Mean(A_{i}^{j} )$
    \If {0$<$Var$<\sigma$ and 0$<$Mean$<\mu$}
    \State $P_{noise}=\Call{crop}{A_{i}^{j}}$
    \State $N_{rw}$=$\Call{Resize}{P_{noise}}$
    \EndIf
\EndFor

\\
\State $\triangleright$ Real-world noise displacement
\Function{N2N-DISPLACEMENT}{$N_{rw}, I_{input}, \alpha$}
    \State $ I_{mix}$$= \alpha \times N_{rw}+(1-\alpha) \times $$I_{input} $
    \State \Return $I_{mix}$
\EndFunction
\\
\State $\triangleright$ Negative augmentation
    \If {$Patch = P_{target}$}
	\State $I_{Neg}$ $= Rot(P_{target}, \beta),$ $\beta \in[0^{ \circ }, 90^{ \circ }, 180^{ \circ }, 270^{ \circ }]$
	\EndIf

\end{algorithmic}
\end{algorithm}

\subsection{Self-supervised Learning with Massive Negatives}
Self-supervised learning~\cite{SSL} helps the model identify key features by creatively designing pseudo-supervised tasks. The negative augmentation module aims to eliminate the problem of sample imbalance in infrared small target detection. As shown in Fig.~\ref{fig:self-supervised}, taking an infrared image with only one target as an example, the negative augmentation module generates four similar unlabeled images. These four images are identical except for the target. The model learns to distinguish features of different targets and backgrounds by analyzing the relationship between different targets in various conditions. In this way, the model can learn more robust feature representations, which are essential for distinguishing targets under real-world conditions. Overall, the combination of self-supervised learning and the negative augmentation module can help the identify key features under different conditions without explicit labels, thereby improving the model's robustness and accuracy.

Given an input infrared image $I_i$, we can enhance it with negative augmentation. Since the negative augmentation makes the sample label $y_i$ remain unchanged. To the $I_i$, we obtain new sample data $I_{Neg}=[I_{Neg}^{1},...,I_{Neg}^{4}]$, along with the label $y_i$. By utilizing these negative samples and their corresponding labels, we can implement self-supervised representation learning. Specifically, we can minimize their features for self-supervised learning. The trained Soft-IoU loss $\mathcal{L}(f_{ \theta }(I_{i}),y_i)$ can be viewed as being optimized as:
\begin{equation}
\mathcal{L}_{Neg} =\mathop{\arg\min}\limits_{\theta}\sum_{i = 0}^{4}\mathcal{L}(f_{ \theta }(I^{i}_{Neg} ),y_{i}),
\end{equation}

where $\mathcal{L}_{Neg}$ epresents the self-supervised loss function. $I_{Neg}^{i}$ refers to the generated image by negative augmentation, $y_{i}$ indicates the corresponding identical label for $I_{Neg}^{i}$. $\mathcal{L}$ employs Soft-IoU loss for computation. By utilizing negative augmentation without requiring additional manual annotations, more diverse features can be learned in a self-supervised fashion. The algorithm of the solving process is shown as Algorithm~\ref{alg:Self-supervised Learning_editing}.

\begin{algorithm}[t!]
\caption{Self-supervised Learning}
\label{alg:Self-supervised Learning_editing}
\begin{algorithmic}

\State \textbf{Input:} $I_{Neg}=[I_{Neg}^{1},I_{Neg}^{2},...,I_{Neg}^{4z}]$: training samples from negative augmentation
\State \textbf {Output:}$Y=[y_1,y_2,...,y_i]$: indicates the corresponding identical label for $I_{Neg}$.
\State
\For{$i\gets1$ to $4z$}
    \State $\mathcal{L}(f_{ \theta }(I_{Neg}^i),y_i)=1-\frac{(\sigma (f_{ \theta }(I_{Neg}^i))\cap  y_i)}{(\sigma (f_{ \theta }(I_{Neg}^i))\cup   y_i)} $
    \If {$\mathcal{L}(f_{ \theta }(I_{Neg}^i),y_i) < \mathcal{L}(f_{ \theta }(I_{Neg}^{i-1}),y_{i-1})$}
    
    \State $\mathcal{L}_{Neg} = \mathcal{L}(f_{ \theta }(I_{Neg}^i),y_i)$
    \EndIf    
\EndFor
\end{algorithmic}
\end{algorithm}

\section{EXPERIMENT}
\subsection{Synthetic SIRST-5K Dataset}
Inspired by other data-scarce fields (such as ship detection and moving vehicle detection), Li et al. developed the NUDT-SIRST~\cite{DNA} dataset for detecting small targets with various target types, sizes, and different clutter backgrounds. However, the amount of data in the NUDT-SIRST  dataset is still far from sufficient. To enhance the quality, quantity, and diversity of the infrared data, we propose a negative sample enhancement strategy to generate a challenging Synthetic SIRST-5K dataset. The training set of the Synthetic SIRST-5K dataset is divided into two subsets: the original set and the development set. The original set includes 663 training images derived from the NUDT-SIRST dataset. The development set contains 4899 images generated using our negative augmentation techniques. On the proposed Synthetic SIRST-5K dataset, both the training set and the development set exhibit similar data distributions yet diverse patterns. We utilize the negative augmentation strategy to enforce the neural model focus on discriminative features of small targets. 

\begin{figure*}[!t]
\centering
\includegraphics[width=0.9\linewidth]{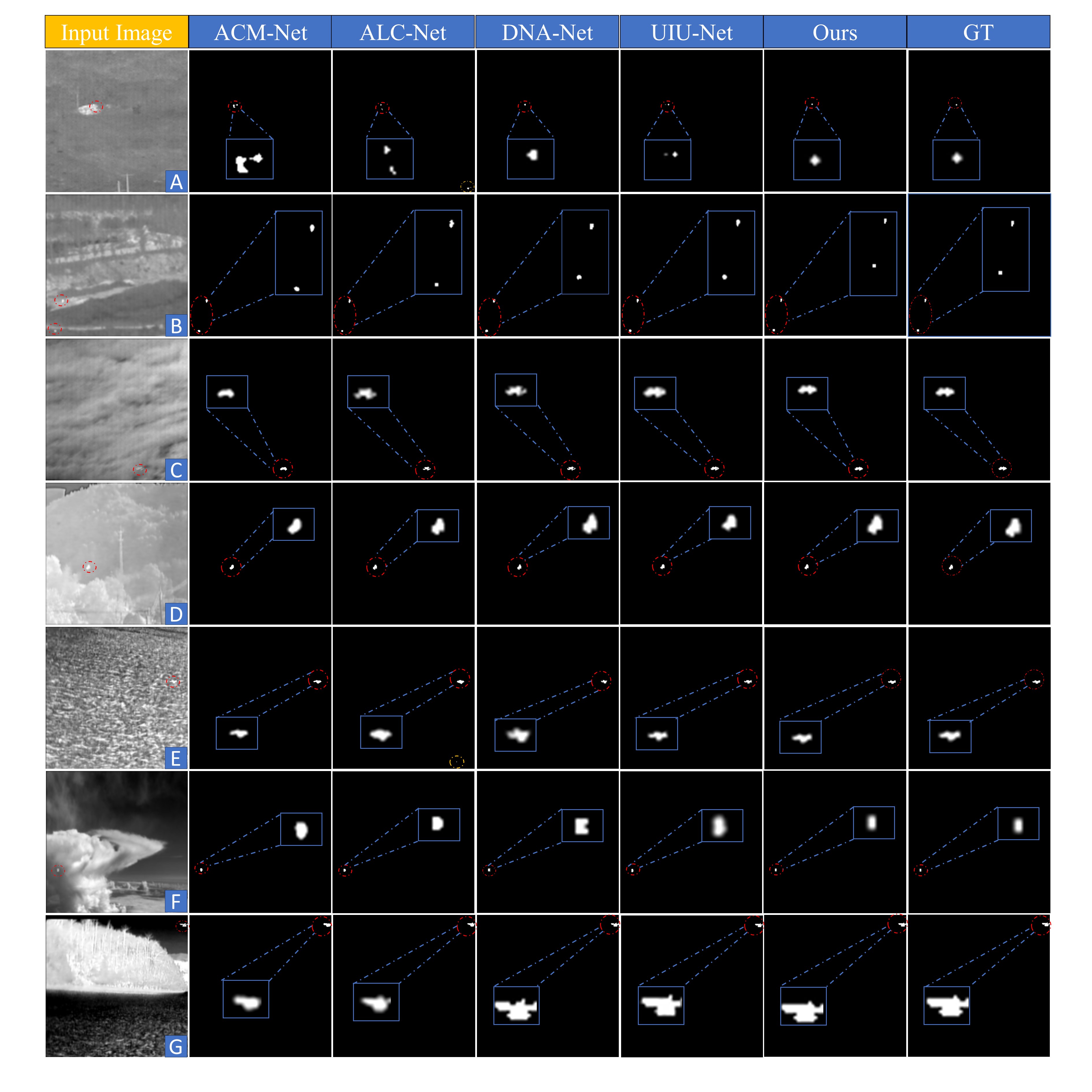}
\caption{The qualitative results obtained from different SIRST detection methods are shown. For better visualization, the small target area is enlarged. Our method performs accurate target localization with a lower false alarm rate.}
\label{fig8}
\end{figure*}

\emph{In our experiments, we utilize the Synthetic SIRST-5K dataset for training, and employ the NUDT-SIRST~\cite{DNA} dataset for evaluation. As shown in Fig.~\ref{fig2} and Fig.~\ref{fig8}, with the same iterations, SIRST-5K significantly improves the robustness of small target detection models, particularly beneficial for detecting targets with a small set of samples.} 

\subsection{Evaluation Metrics}
Traditional target detection mainly utilizes pixel-level evaluation metrics such as Intersection over Union ($IoU$), precision, and recall values. These metrics primarily focus on evaluating the shape of the target. Infrared small targets commonly lack shape and texture information. Therefore, we employ $P_d$ and $F_a$ to evaluate localization accuracy, while using $IoU$ to assess shape accuracy.

\begin{itemize}
    \item[$\bullet$] Intersection over Union: Intersection over Union ($IoU$) is an evaluation metric used in object detection. It is calculated by dividing the area of intersection between the predicted and labeled regions by the area of union between them. The definition is as follows:
    \begin{equation}
    IoU = \frac{|A_{cd}\cap A_{gd} |}{|A_{cd}\cup  A_{gd} |} 
    \end{equation}
    where $A_{cd}$ and $A_{gd}$ respectively represent the predicted region and the labeled region.

    \item[$\bullet$] Probability of Detection: The Probability of Detection ($P_{d}$) is an evaluation metric at the target level. It measures the ratio of the correctly predicted number of targets $T_{correct}$ to the total number of targets $T_{All}$. The definition is as follows:
    \begin{equation}
    P_{d} = \frac{T_{correct}}{T_{All}} 
    \end{equation}
    
    \item[$\bullet$] False-Alarm Rate: False-Alarm Rate ($F_{a}$) is another evaluation metric at the target level. It is used to measure the ratio of falsely predicted pixels to all image pixels $P_{false}$. The definition is as follows:
    \begin{equation}
    F_{a} = \frac{P_{false}}{P_{All}} 
    \end{equation}
\end{itemize}

\begin{center}
    \begin{table*}[ht]
    \caption{ Comparisons with state-of-the-art methods.  The best results are colored in red, and the second-best results are shown in blue. }
    \label{table2}
    \small
    \centering
    \resizebox{1.8\columnwidth}{!}{
    \begin{tabular}{|c|c|c|c|c|}\hline
     \multirow{2}{*}{Type} & \multirow{2}{*}{Method Description}& \multicolumn{3}{c|}{NUDT-SIRST} \\ \cline{3-5} 
     &   & \multicolumn{1}{c|}{$IoU\uparrow (\times 10^2)$ }                                & \multicolumn{1}{c|}{$P_d\uparrow(\times 10^6)$}  & $F_a\downarrow(\times 10^6)$ \\ \hline \hline

    \multirow{2}{*}{Filtering-Based} & Top-Hat~\cite{tophat}  & \multicolumn{1}{c|}{20.72}  & \multicolumn{1}{c|}{78.41}           & 166.7   \\ \cline{2-5}
    
     & Max-Median~\cite{max}  & \multicolumn{1}{c|}{4.197}                 & \multicolumn{1}{c|}{58.41}  & 36.89  \\ \hline
    
    \multirow{2}{*}{Local Contrast-Based} & WSLCM~\cite{WSLCM}    & \multicolumn{1}{c|}{2.283}   & \multicolumn{1}{c|}{56.82}            & 1309 \\ \cline{2-5} 
     & TLLCM~\cite{TLLCM}  & \multicolumn{1}{c|}{2.176}                & \multicolumn{1}{c|}{62.01}  & 1608  \\ \hline
    
     & IPI~\cite{IPI}  & \multicolumn{1}{c|}{17.76}  & \multicolumn{1}{c|}{74.49}               & 41.23 \\ \cline{2-5} 
      & NRAM~\cite{NRAM}   & \multicolumn{1}{c|}{6.927} & \multicolumn{1}{c|}{56.4}             & 19.27\\ \cline{2-5} 
    Low Rank-Based   & RIPT~\cite{RIPT}  & \multicolumn{1}{c|}{29.44}  & \multicolumn{1}{c|}{91.85}           & 344.3  \\ \cline{2-5} 
     & PSTNN ~\cite{PSTNN} & \multicolumn{1}{c|}{14.85}  & \multicolumn{1}{c|}{66.13}              & 44.17  \\ \cline{2-5} 
    & MSLSTIPT~\cite{MSLSTIPT} & \multicolumn{1}{c|}{8.342}                  & \multicolumn{1}{c|}{47.4} & 888.1   \\ \hline
     & MDVSFA-CGAN~\cite{GAN}  & \multicolumn{1}{c|}{75.14} & \multicolumn{1}{c|}{90.47}          & 25.34  \\ \cline{2-5} 
    & ACM-Net~\cite{ACM}  & \multicolumn{1}{c|}{67.08} & \multicolumn{1}{c|}{95.97}            & 10.18   \\ \cline{2-5}
     CNN-Based & ALC-Net~\cite{ALC}  & \multicolumn{1}{c|}{81.4} & \multicolumn{1}{c|}{96.51}  & 9.261 \\  \cline{2-5}
    & DNA-Net~\cite{DNA}  & \multicolumn{1}{c|}{87.09}  & \multicolumn{1}{c|}{ 98.73} & {\color{blue} 4.223} \\ \cline{2-5} 
    &  UIU-Net~\cite{UIU} & \multicolumn{1}{c|}{ \color{blue} {89.00}}    & \multicolumn{1}{c|}{{\color{blue} 98.73}} & 6.021    \\ \cline{2-5} \hline \hline
     \multicolumn{2}{|c|}{ \textbf{Ours}}   & \multicolumn{1}{c|}{\color{red} {\textbf{92.78}}} & \multicolumn{1}{c|}{\color{red} {\textbf{98.84}}} & \color{red} {\textbf{2.735}} \\ \hline
    \end{tabular}
    }
    \label{tab2}
    \end{table*}
\end{center}

\begin{table}[t]
    \caption{Efficiency Analysis.}
    \label{table3}
    \centering
    \small
    \resizebox{0.91\columnwidth}{!}
    {
    \begin{tabular}{|c|c|c|c|c|c|}
    \hline
   Model & ALC-Net & ACM-Net                     & DNA-Net & UIU-Net & Ours                                  \\ \hline \hline
    Params (M)                  & 0.38    &  \textcolor{red}{0.29} & 4.70    & 50.54   & 8.79                                  \\ \hline
    Inference times (S)         & 40.93   &  \textcolor{red}{18.53} &  43.42    & 33.98  & 54.77                                 \\ \hline
    $IoU(\times 10^2)$                   & 81.40   & 67.08                       & 87.09   & 89.00      &  \textcolor{red}{92.78} \\ \hline
    $P_d(\times 10^6)$                    & 96.51   & 95.97                       & 98.73   & 98.73   &  \textcolor{red}{98.84} \\ \hline
    $F_a(\times 10^6)$                    & 9.26   & 10.18                       & 4.22   & 6.02   & \textcolor{red}{2.74}  \\ \hline

    \end{tabular}
    }
    
\end{table}

\subsection{Implementation Details}
In this paper, we choose the U-Net~\cite{unet} model as the backbone for feature extraction. Our model applies 4 downsampling layers and is trained with the Soft-IoU loss function and optimized through the Adagrad~\cite{Adagrad} method and CosineAnnealingLR scheduler. We initialize the weights and biases of the model using the Xavier~\cite{Xavier} method. We set the learn rate, batch size, and epoch size to 0.05, 256, and 1500 respectively, and all models are implemented on a computer equipped with an Nvidia GeForce RTX3090 GPU.

To demonstrate the superiority of our method, we compare our proposed method with state-of-the-art (SOTA) methods. The traditional methods include Top-Hat~\cite{tophat}, Max-Median~\cite{max}, WSLCM~\cite{WSLCM}, TLLCM~\cite{TLLCM}, IPI~\cite{IPI}, NRAM~\cite{NRAM}, RIPT~\cite{RIPT}, PSTNN~\cite{PSTNN}, and MSLSTIPT~\cite{MSLSTIPT}. The CNN-based methods include MDvsFA-cGAN~\cite{GAN}, ACM~\cite{ACM}, ALCNet~\cite{ALC}, DNA-Net~\cite{DNA} and UIU-Net~\cite{UIU}. For a fair comparison, all baseline methods are employed with official implementation.

\subsection{Comparation}

\textbf{Quantitative comparison.} As shown in Tab.~\ref{table2}, traditional methods are usually designed for specific scenes (e.g., specific target size and noise background), and the performances are not good enough in real-world practice. CNN-based methods are limited by the diversity of samples, which all lead to poor prediction results for infrared small targets. Without additional manual annotation, the proposed negative sample enhancement strategy can learn rich features and greater generalization ability. As shown in Tab.~\ref{table2}, compared with all of the state-of-the-art (SOTA) methods, our model shows a significantly improved result. Compared with UIU-Net, our model achieves a 3.78 improvement over $IoU$. Compared with DNA-Net, our model obtains 1.5 point promotion over $F_{a}$, which fully verifies the effectiveness of the proposed negative generation strategy.

\textbf{Qualitative comparison.} The qualitative results of the NUDT-SIRST dataset are shown in Fig.~\ref{fig1} and Fig.~\ref{fig8}. Compared with the current state-of-the-art models, our method has achieved better prediction results. As shown in Fig.~\ref{fig1}, our method can accurately predict the location of small targets with an extremely low false alarm rate. Meanwhile, other CNN-based deep learning networks (such as ACM-Net, ACL-Net) have many false alarms and missed detection regions. As shown in Fig.~\ref{fig8}, the results predicted by our method are more consistent with the ground truth annotation. In conclusion, our method has shown better performance in $P_d$,  $F_a$ and $IoU$ compared to other SOTA methods. This shows that our proposed method can effectively improve the accuracy and robustness of infrared small target detection.

\textbf{Efficiency analysis.} We perform an efficiency analysis with state-of-the-art models in Tab.~\ref{table3}. Compared with other methods, our method shows superior performance and competitive efficiency. Specifically, compared with the UIU-Net method, our method shows better performance in terms of $P_d$, $F_a$ and $IoU$. It is worth noting that the parameters of our model are also far lower than the UIU-Net model parameters. In terms of $P_d$, our model obtains 2 points improvement over ALC-Net and ACM-Net with a similar efficiency. In conclusion, our model achieves the best performance with fewer model parameters and competitive efficiency. The experimental results prove that the proposed model shows higher accuracy with good practicability.

\subsection{Ablation Studies}
To evaluate the effectiveness of the following components: AFF, Noise2Noise displacement, negative augmentation and self-supervised learning. We conducted ablation studies by analyzing each component individually. We used the NUDT-SIRST dataset for training and testing. Meanwhile, the plain model is a model with the same structure as the U-Net. For each part of the ablation study, we rigorously retrained the full model.

\begin{table}[t]
    \caption{Ablation Experiment of Hyperparameter $\alpha$ in Noise2Noise Displacement Module.  }
    \label{table5}
    \centering
    \scriptsize
    \resizebox{1\columnwidth}{!}
    {
    \begin{tabular}{|c|c|c|c|}
    \hline
    Method  Description & $IoU(\times 10^2)$ & $P_d(\times 10^6)$ & $F_a(\times 10^6)$                     \\ \hline
    Plain model    & 86.87   &  97.98 & 3.71   \\ \hline
    $\alpha$ = 0.1  &  \textcolor{red}{90.25} &  98.31 & \textcolor{blue}{2.30} \\ \hline
    $\alpha$ = 0.2     & 88.6    & 98.10     & \textcolor{red}{1.61} \\ \hline
    $\alpha$ = 0.3    & 87.46     & \textcolor{red}{98.94} & 4.60      \\ \hline
    $\alpha$ = 0.4   &  \textcolor{blue}{89.87} & 98.62   & 4.14      \\ \hline
    $\alpha$ = 0.5   & 87.42    &  \textcolor{blue}{98.84} & 8.41   \\ \hline
    \end{tabular}
    }
    \label{tab5}
\end{table}

\begin{table}[t]
    \caption{Ablation Experiment of Hyperparameter $s$ in Negative Augmentation Module.}
    \label{table6}
    \centering
    \scriptsize
    \resizebox{1\columnwidth}{!}
    {
    \begin{tabular}{|c|c|c|c|}
    \hline
    Method  Description & $IoU(\times 10^2)$ & $P_d(\times 10^6)$ & $F_a(\times 10^6)$                     \\ \hline
    baseline    & 90.25   &  98.31 & \textcolor{red}{2.298}   \\ \hline
    $s$ = 3   & \textcolor{red}{92.78}  &  98.84 & \textcolor{blue}{2.735}   \\ \hline
    $s$ = 5   & \textcolor{blue}{91.90}  &\textcolor{blue}{98.94}&  8.732\\ \hline
    $s$ = 7    & 91.79   &  \textcolor{red}{99.05} & 3.585       \\ \hline
    $s$ = 9     & 90.25   &  \textcolor{blue}{98.94} & 16.936       \\ \hline
    
    \end{tabular}
    }
    \label{tab6}
\end{table}

\begin{figure}[!t]
\centering
\includegraphics[width=0.95\columnwidth]{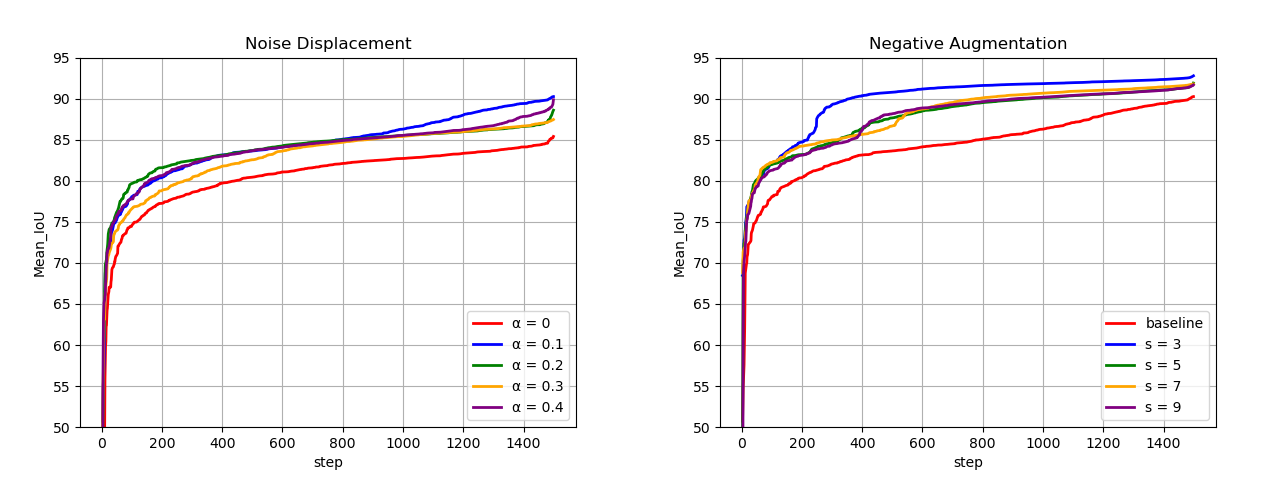}
\caption{Visualization curves of ablation studies for the real-world Noise2Noise Displacement module and Negative Augmentation modules.}
\label{fig9}
\end{figure}

\begin{table}[t]
    \caption{Ablation study of the Noise2Noise Displacement, Negative Augmentation and AFF modules. }
    \label{table4}
    \centering
    \footnotesize
    \resizebox{1\columnwidth}{!}
    {
    \begin{tabular}{|c|c|c|c|}
    \hline
        \thead{Method} &  $IoU\uparrow(\times 10^2)$   & $P_d\uparrow(\times 10^6)$    & $F_a\downarrow(\times 10^6)$ \\ \hline \hline
        \thead{ Plain Model } & 87.11 & 70.05 & 222.61  \\ \hline
        \thead{ w / AFF } & 86.87 & 97.98 & 3.71  \\ \hline
        \thead{ w / AFF + N2N Displacement} & 90.25 & 98.10 & \textcolor{red}{2.30}
        \\ \hline 
        \thead{w / AFF + N2N + Negative Aug. + $\mathcal{L}_{Neg}$ } & \textcolor{red}{92.78} & \textcolor{red}{98.84} & 2.74 \\
        \hline
    \end{tabular}
    }
    \label{tab4}
\end{table}

\textbf{Noise2Noise Displacement.} In this section, we conduct an ablation study on the Noise2Noise displacement. To ensure a fair comparison, we uniformly use the U-Net network model with added AFF for training. As shown in Tab.~\ref{tab5}, the results of the proposed Noise2Noise Displacement are clearly superior to the baseline model. This reveals that the Noise2Noise displacement can improve performance clearly. To perform a comprehensive analysis, we study the noise mixing ratio in Equ.~\ref{eq1}, in which we set $\alpha $ to 0, 0.1, 0.2, 0.3, 0.4, and 0.5, and conduct ablation experiments on them. As shown in Fig~\ref{fig9}, as the noise ratio increases, it is observed that the model performance has a significant decline trajectory. This indicates that too much noise is not conducive to the model training. We found that $\alpha $ = 0.1 has a clear advantage over other values in IoU and $F_a$. Therefore, we choose $\alpha $ = 0.1 as the default parameter for Noise2Noise Displacement.

\textbf{Ablation on Negative Augmentation + $\mathcal{L}_{Neg}$.} We conducted an ablation study on the negative augmentation method to analyze its effectiveness. To ensure a fair comparison, we uniformly used the U-Net network model with added AFF for training. The quantitative results are shown in Tab.~\ref{tab4} and Fig~\ref{fig9}, the results of negative sample enhancement are better than the baseline in terms of $IoU$ and $P_d$. The results show that the negative augmentation strategy has a significant advantage on the model performance. In Section \ref{NA}, we introduce the center flipping for the target point. For the flipping hyperparameter $s$, we take $3 \times 3$, $5 \times 5$, $7 \times 7$, and $9 \times 9$, and perform an further ablation experiment on them.  Compared with other values, $s=3$ has a significant advantage in $IoU$ and $P_d$, therefore we choose $s=3$ as the default value of $s$.

\textbf{Ablation on AFF.} We have improved the effectiveness of the Attention Feature Fusion module. AFF helps the discriminative model focus on target regions by suppressing background regions. As shown in Tab.~\ref{tab4}, the plain model is a basic model with the same structure as the U-Net model. With the AFF mechanism, the proposed algorithm achieves an obvious promotion on $P_d$ and $F_a$.

\section{CONCLUSION AND LIMITATIONS}
In this paper, we proposed a novel negative generation strategy to address the problem of limited data sources in infrared small target detection. Experimental results demonstrate that our method achieves outstanding performance. Compared with other state-of-the-art methods, our method outperforms in $P_d$, $F_a$ and $IoU$. This indicates that our proposed method can effectively improve the accuracy and robustness of infrared small target detection. In summary, the negative sample enhancement strategy provides a new learning paradigm under limited data conditions for research in the fields of autonomous driving, security monitoring, and intelligent transportation systems.

However, the proposed method still has some limitations. It can be seen that our network does not have advantages in terms of the number of parameters and inference times compared with other models. Therefore, reducing the inference times and the number of parameter operations is the focus of our future work while ensuring the performance of the model. In the following work, we will try to use a lighter backbone to achieve breakthroughs in inference time and the number of parameters. Also, We will attempt to further compress the model size by using techniques such as depthwise separable convolution, pruning, and quantization.

\bibliographystyle{IEEEtran}
\bibliography{ref}

\end{document}